\PassOptionsToPackage{numbers,sort&compress}{natbib}

\documentclass{article}

\usepackage[preprint]{neurips_2025}

\usepackage[utf8]{inputenc} 
\usepackage[T1]{fontenc}    
\usepackage{hyperref}       
\usepackage{url}            
\usepackage{booktabs}       
\usepackage{amsfonts}       
\usepackage{mathrsfs}       
\usepackage{nicefrac}       
\usepackage{microtype}      
\usepackage{xcolor}         
\usepackage{graphicx,subcaption} 
\usepackage{tikz}
\usepackage{bbm}
\usepackage{amsmath} 
\usepackage{float}
\usepackage{placeins}
\usepackage{chngcntr}

\title{Beyond Conventional Transformers: The Medical X-ray Attention (MXA) Block for Improved Multi-Label Diagnosis Using Knowledge Distillation}

\author{%
  Amit Rand\thanks{Equal contribution.} \\
  Department of Mathematics\\
  University of California, Los Angeles\\
  Los Angeles, CA 90095 \\
  \texttt{amit.rand@ucla.edu} \\
  \And
  Hadi Ibrahim\footnotemark[1] \\
  Department of Mathematics\\
  University of California, Los Angeles\\
  Los Angeles, CA 90095 \\
  \texttt{{hadiibrahim@ucla.edu}} \\
}

\begin{document}

\maketitle

\begin{abstract}
  Medical imaging, particularly X-ray analysis, often involves detecting multiple conditions simultaneously within a single scan, making multi-label classification crucial for real-world clinical applications. In this paper, we present the Medical X-ray Attention (MXA) block, a novel attention mechanism tailored specifically to address the unique challenges of X-ray abnormality detection. The MXA block enhances traditional Multi-Head Self Attention (MHSA) by integrating a specialized module that efficiently captures both detailed local information and broader global context. To the best of our knowledge, this is the first work to propose a task-specific attention mechanism for diagnosing chest X-rays, as well as to attempt multi-label classification using an Efficient Vision Transformer (EfficientViT). By embedding the MXA block within the EfficientViT architecture and employing knowledge distillation, our proposed model significantly improves performance on the CheXpert dataset, a widely used benchmark for multi-label chest X-ray abnormality detection, with data augmentation methods applied. Our approach achieves an area under the curve (AUC) of 0.85, an absolute improvement of 0.19 compared to our baseline model’s AUC of 0.66, corresponding to a substantial approximate 233\% relative improvement above random guessing (AUC = 0.5).
  \footnotemark[1]\footnotetext[1]{The code and results for this study are available at: \href{ https://github.com/Hadi-M-Ibrahim/Beyond-Conventional-Transformers/.}{https://github.com/Hadi-M-Ibrahim/Beyond-Conventional-Transformers}.}
\end{abstract}


\section{Introduction}
\label{sec:intro}
Chest X‑rays (CXR) are the most common medical images, yet one scan often reveals several co‑existing pathologies that automated readers must detect simultaneously.  Vision Transformers (ViTs) \citep{dosovitskiy2020image} excel at global reasoning, but their quadratic self‑attention makes deployment on high‑resolution CXR costly in both parameters and floating‑point operations (FLOPs).  Pure convolutional networks, e.g.\ DenseNet‑121 \citep{rajpurkar2017chexnet}, run fast but miss long‑range cues spread across the thorax.  The clinical setting therefore demands models that (i) learn multi‑label dependencies, (ii) highlight subtle local findings, and (iii) fit the limited compute of hospital workstations.

Recent efficient transformers, DeiT \citep{touvron2021training}, Swin \citep{liu2021swin}, EfficientViT \citep{liu2022efficientvit}, trim FLOPs via knowledge distillation (KD) or windowed attention, yet none are specialized for medical images where abnormalities occupy just a few pixels. Moreover, public CXR datasets such as CheXpert \citep{irvin2019chexpert} remain modest by ViT standards, amplifying over‑fitting and class imbalance.  We address these gaps with a domain‑aware transformer that pairs global self‑attention with inexpensive region attention and learns from a calibrated CNN teacher.

\paragraph{Contributions.}
Our work makes three principal contributions: (i) We extend the EfficientViT to multi-label chest-X-ray diagnosis by replacing the cross-entropy objective with a class-balanced BCE with logit loss, enabling a single forward pass to predict all 14 CheXpert labels. (ii) We introduce the lightweight Medical X-ray Attention (\textbf{MXA}) block, which pairs dynamic ROI pooling~\citep{girshick2015fast} with channel–spatial gating~\citep{woo2018cbam} in parallel to multi-head self-attention, capturing subtle lesions without increasing asymptotic complexity. (iii) We devise a multi-label knowledge-distillation scheme in which a frozen DenseNet-121 teacher provides soft targets, guiding the transformer on rare pathologies and stabilizing training on the 224 k-image CheXpert corpus~\citep{cohen2021torchxrayvisionlibrarychestxray}.

On CheXpert our EfficientViT-MultiLabel-M5 model reaches 0.85 mean ROC‑AUC, improving the M5 EfficientViT baseline by +0.19 while preserving its 5.7G FLOPs footprint. MXA alone delivers half the gain; distillation adds the remainder, confirming their complementary roles.  The resulting architecture offers an attractive accuracy‑to‑efficiency trade‑off for point‑of‑care deployment and illustrates how task‑specific attention bridges the gap between generic ViTs and radiologist needs.

\section{Related Works}
\paragraph{Multi‑label image classification and CheXNet.} Deep CNNs, from AlexNet to EfficientNet, drove single-label ImageNet accuracy upward, but Vision Transformers (ViTs) surpassed them once large-scale data or distillation became available \citep{krizhevsky2012imagenet,he2016deep,tan2019efficientnet,dosovitskiy2020image,touvron2021training}.
In multi-label chest X-ray diagnosis several findings (e.g., consolidation, edema, cardiomegaly) often co-occur. The per-label sigmoid activation followed by Binary Cross-Entropy (BCE) loss, widely adopted for such tasks, optimizes each label in isolation. This independence assumption disregards inter-label structure and tends to underweight infrequent diseases, leading to poorer AUC on rare classes \citep{wang2016cnn,huynh2020joint}.
CheXNet mitigated class imbalance by fine-tuning DenseNet-121 on 112 k NIH scans and achieved radiologist-level pneumonia detection \citep{rajpurkar2017chexnet}; however, its purely convolutional backbone still treated pathologies independently and lost accuracy on external cohorts. Transformer variants such as TransMed explicitly model label dependencies, yet their quadratic self-attention inflates FLOPs, limiting bedside deployment \citep{shamshad2023transmed}.
Our approach grafts a lightweight Medical X-ray Attention block onto EfficientViT \citep{liu2022efficientvit} and distills knowledge from a calibrated DenseNet teacher, preserving CheXNet’s clinical strengths while delivering ViT-level global reasoning at workstation-level cost.

\paragraph{CheXpert benchmark.} CheXpert offers 224k radiographs from 65k patients with rule‑mined, uncertainty‑annotated labels for the same 14 findings, providing a stronger test bed than NIH for real‑world deployment \citep{irvin2019chexpert}.  Its public leaderboard has steered advances ranging from label‑uncertainty strategies (e.g., U‑on‑1, self‑training) to loss designs that directly maximize AUC on imbalanced data \citep{yuan2021large}.  Transformer backbones, including DeiT and Swin, now dominate the leaderboard but often rely on extensive augmentations and multi‑crop inference that inflate latency. However, computationally heavy models impede clinical application where GPUs may not be available.  We thus adopt CheXpert as our primary benchmark yet target an efficiency regime suitable for clinical integration: EfficientViT halves FLOPs versus Swin while our MXA block and distillation recover the accuracy gap.  By evaluating under identical U‑1 labeling and augmentation settings, we isolate the benefit of task‑specific attention rather than dataset heuristics.

\paragraph{DenseNet‑121 as teacher.} DenseNet‑121’s dense skip connections deliver strong feature reuse with only 8M parameters \citep{huang2017densely}; consequently it is the default backbone in many open‑source chest‑X‑ray libraries such as TorchXRayVision.  When pretrained on CheXpert the model attains robust AUC across diverse pathologies and exhibits calibrated probability outputs, making it an ideal teacher for KD.  Prior student–teacher pairs have compressed DenseNet into MobileNet or ShuffleNet for on‑device screening, but rarely into transformers \cite{li2020covidmobilexpert, zhang2018shufflenet}. We keep the teacher frozen and distill its logits into an EfficientViT student, showing that soft multi‑label distillation narrows the performance gap while letting the transformer exploit global context unavailable to the convolutional teacher.

\section{Methods}
\subsection{Multi-labeling with Efficient Vision Transformers}
\label{sec:multilabel_efficientvit}

While Cream's EfficientViT\footnote{\url{https://github.com/microsoft/Cream/tree/main/EfficientViT}} was engineered for single‑label tasks (e.g. ImageNet classification) \cite{liu2022efficientvit}, CXR diagnosis demands multi‑label predictions, as several abnormalities can co‑occur. To handle the co-occurrence of multiple pathologies in a single chest X-ray, we adapt EfficientViT’s single-label framework by redesigning its output head and training loss for multi-label prediction.  Specifically, we replace the softmax head with a per-class linear layer trained under a class-balanced binary cross-entropy objective, enabling one efficient transformer pass to simultaneously predict all target abnormalities.

\paragraph{Label representation \& loss objective.}
We follow the “U‑1’’ protocol (Appendix~\ref{app:u1_formal}), motivated by best practice ~\cite{irvin2019chexpert}~\cite{phang2019adjusting}. EfficientViT’s final pooling layer is replaced by a $(B,C)$ linear head whose logits $o=(o_1,\dots,o_C)$ are trained with Binary Cross‑Entropy with Logits (BCEWL); full definition in Appendix~\ref{app:bcewl}. The formulation naturally accommodates independent per‑pathology predictions, while letting the network exploit both definite and uncertain positives, and yields probability scores $\sigma(o_c)\!\in\![0,1]$ used for AUC and threshold accuracy metrics. 

\paragraph{Training, Loss \& Evaluation.} We train our EfficientViT student on batches of size \(B\) by minimizing a convex combination of the ground-truth multi-label BCEWL and a soft-label knowledge-distillation term from our frozen DenseNet-121 teacher. Specifically, with student logits \(O^s\in\mathbb{R}^{B\times 14}\), true labels \(Y\in[0,1]^{B\times14}\) under the U-1 protocol (Appendix \ref{app:u1_formal}), and teacher probabilities \(p^t=\sigma(O^t)\), our total loss is
\[
\mathscr{L}_{\text{total}}
\;=\;
(1-\alpha)\,
\mathscr{L}_{\text{BCEWL}}\!\bigl(O^{s},Y\bigr)
\;+\;
\alpha\,
\mathscr{L}^{\text{soft}}_{\text{KD}}\!\bigl(O^{s},p^{t}\bigr),
\]
where $\alpha\!\in\![0,1]$ controls the distillation strength. During each forward pass we compute \(O^s=\mathrm{EfficientViT}(\text{samples})\) and back-propagate through the student only. At validation time we evaluate the exact same loss and additionally measure two key metrics: (1) mean area under the ROC curve (AUC) across all 14 pathologies, which captures discrimination across thresholds, and (2) per-label accuracy at a 0.5 probability cutoff, as a proxy for binary clinical decision performance, by thresholding \(\sigma(O^s)\) and comparing to \(Y\). All reported metrics are averaged over three random seeds, with 95\% confidence intervals where applicable. Our primary objective is to boost mean AUC over the vanilla EfficientViT baseline, especially on under-represented diseases, while maintaining or exceeding baseline threshold accuracy.

\subsection{Medical X-ray Attention (MXA) Block}
We propose a new block designed to enhance the efficiency and accuracy of transformer-based architectures in multi-label clinical X-ray abnormality detection and diagnosis. The MXA block is composed of Dynamic Region of Interest (ROI) \cite{girshick2015fast} Pooling and Convolutional Block Attention Module (CBAM)-Style Attention \cite{woo2018cbam}, each of which focuses on improving localized model performance. We integrate this block in parallel with the Multi-Head Self-Attention (MHSA) layers within EfficientViTs \cite{liu2022efficientvit} to optimize both computational efficiency and accuracy.

\begin{figure}[h!]
    \centering
    \includegraphics[width=\textwidth]{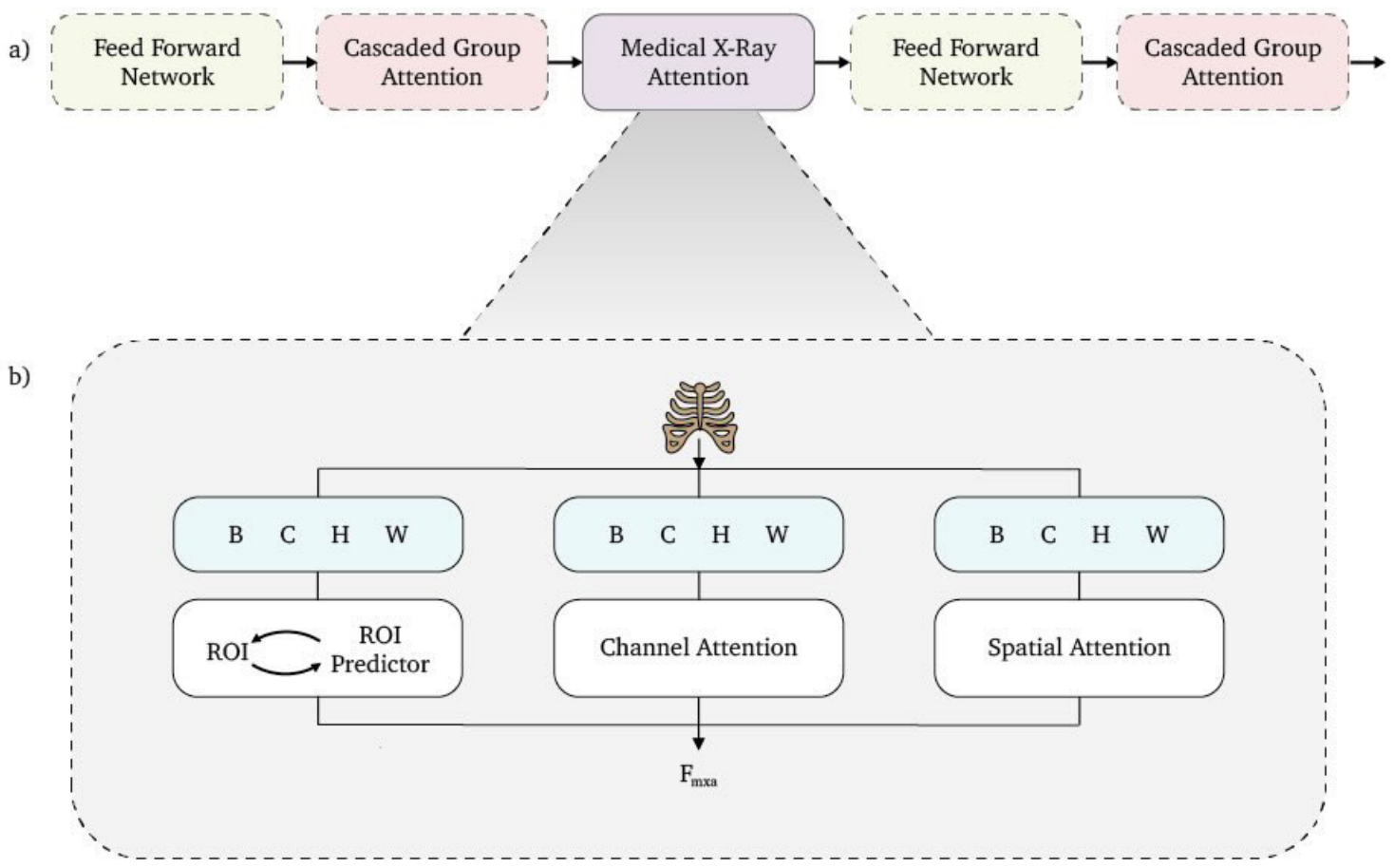}
    \caption{Integration of the Medical X-ray Attention (MXA) Block Architecture. The MXA block injects ROI pooling + CBAM gating in parallel with MHSA, focusing compute on abnormal regions.}
    \label{fig:mxa_block_architecture}
\end{figure}

\paragraph{Dynamic region-of-interest (ROI) pooling.} 
Multi-label X-ray classification often includes multiple abnormal regions and purely global feature pooling might dilute small or subtle findings. To identify the most relevant regions for targeted analysis, consequently reducing computational overhead, we propose a Dynamic ROI pooling approach. By dynamically pooling region-of-interest (ROI) features, the model sub-sectionalizes critical areas while ignoring extraneous regions. 

We employ a learnable ROI-pooling module in which a lightweight convolutional predictor generates normalized bounding-box coordinates
$\mathrm{ROI}_i = [x_1, y_1, x_2, y_2]$ for each feature map $F\in\mathbb{R}^{B\times C\times H\times W}$. These boxes are used to crop and rescale $F$, and the multi-label classification loss back-propagates through the ROI predictor so that box localization and classification parameters are optimized end-to-end (see Appendix~\ref{app:def_roi} for full details). After cropping the predicted ROI from the feature map, we resize it to the original feature‑map size via bilinear interpolation:
\[
F^{(i)}_{\text{pooled}}
\;=\;
\operatorname{Resize}\!\bigl(F_{\mathrm{ROI}_i},\, (H,W)\bigr),
\]
where $F^{(i)}_{\text{pooled}}$ is the pooled ROI feature map and
$\operatorname{Resize}(\cdot)$ denotes bilinear interpolation. By pooling only the relevant region to full resolution, we reduce the overhead that would otherwise come from processing the entire X-ray at high granularity \cite{shen2019roi}. This ensures compatibility with subsequent operations while focusing on the most critical regions. We use a kernel size of 3 in certain layers (e.g. local window attention or ROI-based convolutional components) as a practical middle ground for ROI between capturing sufficient local context and avoiding over-fitting in CXR data. Prior work in medical image analysis (e.g., \cite{ronneberger2015unet}) has similarly found that intermediate kernel sizes strike a good balance between detail preservation and robust feature extraction.

\paragraph{Convolutional Block Attention Module (CBAM)‑style attention.}

To further refine the representation and enhance localized feature learning, we incorporate a CBAM \cite{woo2018cbam} into the MXA block. CBAM sequentially applies channel attention and spatial attention to the pooled feature maps, enabling the network to focus on the most informative channels and spatial regions. This combined attention helps the network prioritize more pathologically relevant features, crucial in medical images where anomalies can be small or faint. 

Channel attention determines which feature channels contribute the most to the classification task. This is achieved by globally aggregating the feature map along the spatial dimensions using global average pooling (GAP) and global max pooling (GMP), followed by a shared multi‑layer perceptron (MLP) \cite{lin2013network}; see Appendix \ref{app:cbam_attention} for the formal definition.

The combined effect of channel and spatial attention refines the feature representation by prioritizing both the most informative channels and regions. Incorporating CBAM into the MXA block thus provides a richer and more adaptive representation of X‑ray feature maps. 

\paragraph{Parallel integration with MHSA.} 

To further enhance feature representation and ensure that the model effectively
utilizes both global dependencies (captured by MHSA \cite{vaswani2017attention})
and localized features (captured by MXA), we propose integrating the MXA block
in parallel with the MHSA layers in EfficientViTs.
\[
F_{\text{output}}
\;=\;
F_{\text{MHSA}}
\;+\;
F_{\text{MXA}},
\]
where $F_{\text{MHSA}}$ represents the output from the MHSA layer, and
$F_{\text{MXA}}$ is the refined feature map from the MXA block. Refer to
Appendix~\ref{app:mxa_qual}, Fig.~\ref{fig:mxa_demo} for an illustration of the MXA block and its integration into the overall transformer framework.

\subsection{Knowledge Distillation}
\label{subsec:kd}
\paragraph{Transformation layer and label mapping.} Knowledge distillation (KD) transfers knowledge from a large, well-trained teacher to a compact student model \cite{hinton2015distilling}.
Within our EfficientViT framework for multi-label CXR classification, we use the TorchXRayVision\footnote{\url{https://github.com/mlmed/torchxrayvision}} DenseNet-121 teacher (specifically the ``densenet121-res224-chex'' variant) \cite{huang2017densely, rajpurkar2017chexnet, cohen2020limits}, pretrained on CheXpert \cite{irvin2019chexpert} and producing 18 logits.  Our student predicts 14 pathologies, so we insert a lightweight adapter to align the teacher’s 18‑dimensional output with the 14 labels expected by the
student. Let $O^{\text{t}}\in\mathbb{R}^{18}$ denote the teacher logits.  
First, a fixed index map $M$ permutes the logits so that each student index
$k$ receives the correct teacher value (e.g., $M(1)=17$).  
Second, because the student includes a \emph{No Finding} class that is absent in the teacher, we synthesize its probability by combining the teacher’s per‑pathology scores:
\[
p_{\textsc{NF}}
   = \prod_{i=1}^{18}\bigl(1-\sigma(O^{\text{t}}_{i})\bigr),
\qquad
\text{logit}_{\textsc{NF}}=\sigma^{-1}(p_{\textsc{NF}}).
\]
This quantity captures the likelihood that no abnormality is present. Finally, logits corresponding to teacher labels with no student counterpart  (e.g., Support Devices, Pleural Other) are set to zero so they receive no gradient. This procedure reconciles label-space mismatches and passes a clean 14-dimensional target to the student, allowing KD to sharpen recognition of subtle or co-occurring findings.

\paragraph{Soft distillation and dynamic weighting.}
The adapted teacher probabilities from the original  $O^{\text{t}}\in\mathbb{R}^{18}$ logits are represented as \(p^{\mathrm{t}}=\sigma(O^{\mathrm{t}})\). We adopt \emph{soft} distillation because it preserves the teacher’s uncertainty. To focus learning where the teacher is least confident, we introduce dynamic label weights
\[
w_{j}
\;=\;
1-\frac{1}{B}\sum_{i=1}^{B}p^{\mathrm{t}}_{ij},
\]
where \(B\) is the batch size and \(p^{\mathrm{t}}_{ij}\) the teacher probability for label \(j\) on sample \(i\).  
These weights down-scale loss contributions from high-confidence teacher predictions \((p^{\mathrm{t}}_{ij}\!\approx\!1)\) and amplify ambiguous cases; refer to Appendix \ref{app:distill_term} for the rigorous soft distillation term. 

\section{Experiments}
We evaluate our multi-label M5 EfficientViT for CXR classification, comparing a naive multi-label EfficientViT baseline against our proposed method that integrates KD and the MXA block. In both cases, models are trained and validated on the CheXpert dataset \cite{irvin2019chexpert}, and performance is reported via standard multi-label metrics (average accuracy, F1-score, and micro averaged AUC) at a positive label threshold of 0.5.

\subsection{Implementation Details}
\label{subsec:implementation_details}
The EfficientViT-MultiLabel-M5 is implemented in \texttt{PyTorch 2.5.1} \cite{paszke2019pytorch} using the \texttt{timm 0.5.4} library \cite{rw2019timm}. Training is performed from scratch for 50~epochs on a single NVIDIA~H100 GPU, employing the AdamW optimizer \cite{loshchilov2018adamw} with a weight decay of 0.025 and the ReLU activation function within all EfficientViT blocks. We use the preset train–validation split provided by CheXpert, ensuring consistency with prior benchmarks. Chest X‑ray images are resized to $224\times224$, and a $16\times16$ patch size is used for tokenization. The \texttt{LocalWindowAttention} module within each M5 EfficientViT block employs a $7\times7$ local window, balancing local context capture with computational efficiency.

We train with a batch size of~512 and set an initial learning rate of
$1\times10^{-3}$, following a cosine scheduler with a minimum of
$1\times10^{-5}$. Gradient clipping is applied with a maximum norm of 0.02
using adaptive gradient clipping (AGC). The learning‑rate warm‑up lasts for
5 epochs; cosine decay (rate 0.1) spans 30 decay epochs and is followed by
10 cool‑down epochs. EMA (exponential moving average) is enabled with a decay
factor of 0.99996.

For KD we employ DenseNet‑121 \cite{huang2017densely} (``densenet121‑res224‑chex'' variant \cite{cohen2020limits}) pretrained on CheXpert. Its final classification layer is removed, the network is fine‑tuned for multi‑label classification, and kept frozen during student training. Soft distillation is applied with $\alpha=0.5$ and temperature $\tau=1.0$. Our EfficientViT-MultiLabel-M5 backbone (full spec in Appendix A, Table~A.1) outputs 14 logits (one per pathology). Bi-cubic interpolation is used for resizing, with data augmentation via RandAugment (``rand‑m5‑mstd0.2‑inc2''). Stochastic depth, dropout, mixup, and cutmix are disabled for stable training. EfficientViT-MultiLabel-M5 uses stage widths $\{192,288,384\}$, depths $\{1,3,4\}$, and heads $\{3,3,4\}$.

\subsection{Training Protocol}
\paragraph{Baseline (EfficientViT).}
We train the three‑stage M5 EfficientViT backbone with the AdamW optimizer and a cosine learning‑rate schedule initialized at
\(1\times10^{-3}\). The objective is the sum of \(\operatorname{BCEWithLogitsLoss}\) over the 14 thoracic findings. No KD) or MXA modules are included, so the network reduces to a straightforward multi‑label EfficientViT with patch embeddings and cascaded group attention.
\paragraph{Proposed (EfficientViT+MXA+KD).}
Each stage now contains a parallel MXA block that learns dynamic regions of
interest and applies CBAM‑style channel–spatial attention to emphasize
critical areas. A DenseNet‑121 pretrained on CheXpert (frozen) acts as teacher, while the student minimizes a weighted sum of the ground‑truth \(\operatorname{BCEWithLogitsLoss}\) and the KD term \(\mathcal{L}_{\text{distill}}\) computed from teacher probabilities. An ROI predictor is trained end‑to‑end, producing bounding boxes that are bilinearly up sampled before further attention processing. The optimizer and learning‑rate schedule match the baseline, with updates applied only to student parameters.

\section{Results}
We evaluate and report our main findings on the CheXpert dataset, comparing the proposed M5 EfficientViT with the MXA Block and KD approach against a baseline multi-label M5 EfficientViT. We further present an ablation study in which different components (the MXA block and KD) are incrementally introduced. Finally, we analyze the per-pathology performance to highlight the strengths and weaknesses of our approach for specific conditions.

\begin{table}[h!]
\centering
\caption{Overall Performance Comparison}
\label{tab:overall_performance}
\begin{tabular}{lcccc}
    \toprule
    \textbf{Model} & \textbf{Accuracy (\%)} & \textbf{Loss} & \textbf{F1-Score} & \textbf{AUC} \\
    \midrule
    Baseline (M5 EfficientViT) & 84.0 & 0.679 & 0.476 & 0.661 \\
    Proposed (M5 EfficientViT + MXA + KD) & 84.4 & 0.406 & 0.599 & 0.8529 \\
    \bottomrule
\end{tabular}
\end{table}

\subsection{Overall Performance}

\begin{figure}[h!]
  \centering
  \includegraphics[width=0.8\linewidth]{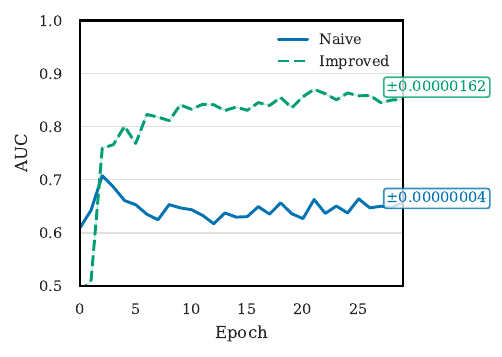}
  \caption{Training AUC over 30 epochs for baseline and proposed models, showing stable convergence and a consistent performance margin. The improved model converges faster and sustains 0.19 higher AUC than the naive baseline throughout training, evidencing durable gains.}
  \label{fig:train_curve}
\end{figure}

Table~\ref{tab:overall_performance} compares the baseline multi‑label M5
EfficientViT model with our proposed method, which integrates the MXA block and
KD across 50~epochs on validation. The baseline achieves an
ROC AUC of $0.66$, whereas our proposed method attains an ROC AUC of $0.85$.
Notably, this improvement of $0.19$ in absolute terms ($66\%\to85\%$), 
a relative gain of roughly $29\%$, corresponds to a large margin over the
random‑guessing baseline of $50\%$ AUC. We also observe consistent improvements
in the average accuracy, F$_1$‑score, and cross‑entropy loss, demonstrating that
our model captures both robust global signals (via the transformer backbone)
and subtle local anomalies (through MXA). Figure~\ref{fig:train_curve} presents AUC (mean $\pm1\sigma$) per epoch over three runs. The proposed model climbs above $0.8$ by epoch~10, compared to $0.65$ for the baseline, demonstrating faster early learning. Both curves plateau by epoch~30 with $\sigma<1.2\times10^{-6}$, indicating stable convergence. A consistent $\approx0.19$ micro-AUC gap persists throughout training, evidencing robust, enduring gains from MXA and KD.

\subsection{Ablation Study} 

To isolate the contribution of each module we train three variants, each for
50~epochs with identical, strengthened augmentations: (i) U1 is the multi‑label M5 EfficientViT that follows the U‑1
label protocol but omits MXA and KD, (ii) U1+MXA adds the Medical X‑ray Attention block in parallel with the transformer attention, and (iii) U1+MXA+KD further incorporates KD from a frozen DenseNet‑121 teacher.

From Table~\ref{tab:ablation_results}, we observe that adding the MXA block (U1 + MXA) reduces the total loss by almost half compared to the baseline and raises the micro-AUC from 0.666 to 0.834. This indicates the ROI-based feature extraction and CBAM-style attention significantly enhance local discrimination. With both MXA and KD (U1 + MXA + KD), we see a further jump to a micro-AUC of 0.839. While the gain is smaller than MXA’s jump, it suggests that teacher-driven soft labels help the student model disambiguate harder pathologies.

\begin{table}[h!]
\centering
\caption{Ablation Study Comparison of M5 Models Overall Across 50 Epochs}
\label{tab:ablation_results}
\begin{tabular}{lc}
    \toprule
    \textbf{Augmentation} & \textbf{AUC} \\
    \midrule
    U1 + Better Augs & 0.6659 \\
    U1 + Better Augs + MXA & 0.8344 \\
    U1 + Better Augs + MXA + KD & 0.8393 \\
    \bottomrule
\end{tabular}
\end{table}

\begin{figure}[h!]
  \centering
  \begin{subfigure}[t]{0.48\textwidth}
    \includegraphics[width=\linewidth]{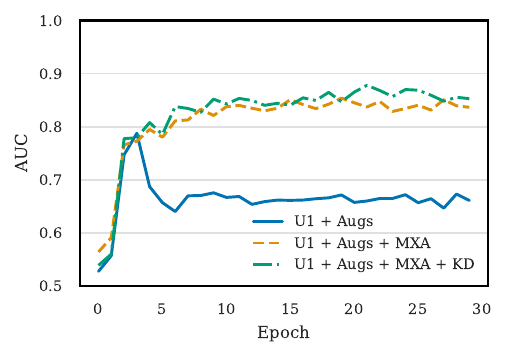}
    \caption{Training AUC curves}
    \label{fig:ablation_curve}
  \end{subfigure}
  \hfill
  \begin{subfigure}[t]{0.48\textwidth}
    \includegraphics[width=\linewidth]{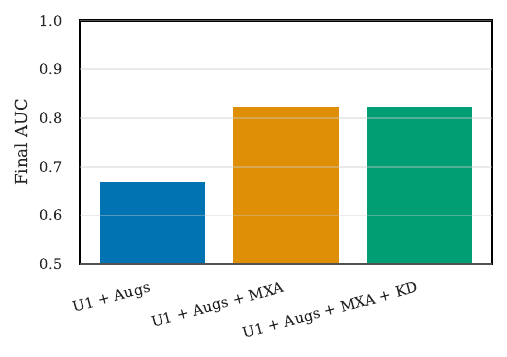}
    \caption{Final-epoch AUC bars}
    \label{fig:ablation_bar}
  \end{subfigure}
  \caption{(a) Training AUC over 30 epochs. (b) Final-epoch AUC for each ablation. MXA alone delivers the largest jump in AUC; adding KD yields an additional boost, lifting performance from $0.66 \rightarrow 0.85$}
  \label{fig:combined_ablation}
\end{figure}

\subsection{Per-Pathology Analysis} To gain further insight into the model performance across individual pathologies, Table~\ref{tab:ablation_pathology_auc} reports the per-pathology AUC values for the same three configurations in the ablation study. 

Our proposed method demonstrates consistent improvements across most conditions, with rare pathologies exhibiting the most significant gains. Categories such as Enlarged Cardiomediastinum (ECM) and Consolidation (CON) see considerable gains once MXA and KD are introduced, reflecting the model’s improved sensitivity to subtle indicators. The dynamic label weighting and transformation layer help mitigate the impact of uncertain or noisy labels, leading to more robust predictions. Pathologies like No Finding (NF) or Lung Opacity (LO) rely more on broader context, whereas localized labels such as Lung Lesion (LL) appear more challenging. LL’s AUC is less stable across ablations, confirming that localized pathologies remain tough even with ROI-based attention. For Fracture (FX), the data often lack strong ground-truth signals, resulting in NaN or missing AUC in all ablation settings. This underscores the dataset’s inherent limitations and the reliance on uncertain label mapping. 

Overall, the per-pathology results confirm that both global context and local enhancement matter for CXR classification. Our final approach (U1 + MXA + KD) achieves the highest AUC on the majority of pathologies, especially for those that had been historically underrepresented or more difficult to detect. In practice, we find that integrating MXA yields more interpretable attention maps, often localizing subtle regions (e.g., small lesions) that the baseline missed. Combined with KD, the model better distinguishes confounding signs such as cardiomegaly vs. consolidated lung fields, leading to fewer false positives in real-world scenarios.

\begin{table}[h!]
\centering
\caption{Ablation Study Per-pathology AUC Comparison Across 50 Epochs}
\label{tab:ablation_pathology_auc}
\begin{tabular}{lccc}
    \toprule
    \textbf{Pathology} & \textbf{U1 + Better Augs} & \textbf{U1 + Better Augs + MXA} & \textbf{U1 + Better Augs + MXA + KD} \\
    \midrule
    NF  & 0.50 & 0.85 & 0.82 \\
    ECM & 0.50 & 0.50 & 0.53 \\
    CM  & 0.50 & 0.50 & 0.69 \\
    LO  & 0.82 & 0.82 & 0.85 \\
    LL  & 0.50 & 0.50 & 0.32 \\
    ED  & 0.78 & 0.86 & 0.87 \\
    CON & 0.50 & 0.77 & 0.81 \\
    PNA & 0.50 & 0.50 & 0.75 \\
    ATL & 0.50 & 0.78 & 0.77 \\
    PTX & 0.50 & 0.81 & 0.79 \\
    PE  & 0.82 & 0.91 & 0.91 \\
    PO  & 0.50 & 0.99 & 0.65 \\
    FX  & NaN   & NaN   & NaN   \\
    SD  & 0.77 & 0.81 & 0.82 \\
    \bottomrule
\end{tabular}
\end{table}

\begin{figure}[h!]
  \centering
  \includegraphics[width=\linewidth]{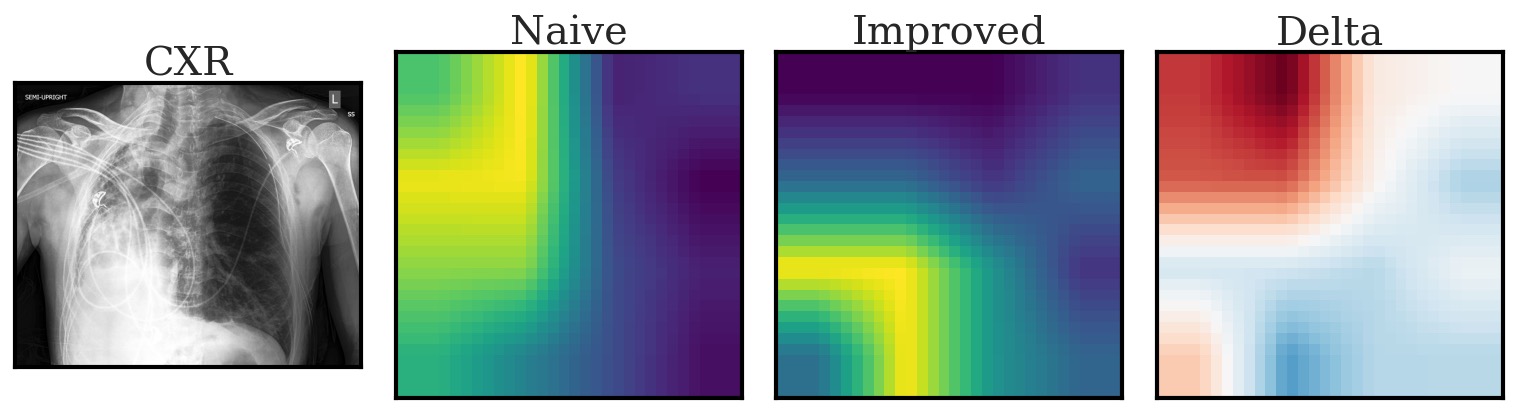}
  \caption{After 25 training epochs, an inference pass of the improved model with the MXA yields more focused and clinically meaningful attention on a CXR with pneumonia. Each heat-map pixel is the normalized attention score for that image patch. Bright yellow in the Naive/Improved panels = high attention; dark purple = low attention. The Delta panel shows the difference in attention: red indicates regions where MXA attends \textit{less} than naive MHSA, blue where it attends \textit{more}, and white means no change. MXA suppresses spurious focus on the shoulders while amplifying attention over the lower left lung field where consolidation is visible, mirroring radiologist practice.}
  \label{fig:qualitative_mxa}
\end{figure}

\section{Discussion}
\label{sec:dis}
This paper demonstrates the efficacy of augmenting general-purpose attention mechanisms with task-specific modules tailored to the unique challenges of a given domain. By introducing the Medical X-ray Attention (MXA) block and leveraging KD, we address critical limitations in training models for medical imaging classification, such as the need for computational efficiency, subtle feature extraction, and multi-label classification. The integration of these techniques into the Efficient Vision Transformer (EfficientViT) architecture highlights the broader potential of task-specific attention mechanisms and resource-efficient training strategies for advancing AI across diverse domains.

\subsection{Limitations}
\label{subsec:limitations}
Despite its promising results, our approach has several limitations. First, the reliance on CheXpert introduces biases inherent to the dataset, such as label uncertainty and imbalance. While our U-1 label-handling scheme and refined augmentations mitigate some of these issues, further steps are often warranted to address both the social bias \cite{lee2023survey} and the hidden stratification \cite{oakden2020hidden} of any given medical data set. Additionally, the MXA block’s reliance on ROI predictions may face challenges with pathologies that lack localized features, such as fractures, which can occur anywhere in the body and are associated with relatively low AUC scores on validation runs.

Furthermore, our study was constrained by limited computational resources, which introduced additional challenges. Specifically, we were unable to generate a custom teacher model, a step that we hypothesize could have significantly enhanced the overall performance of KD. The lack of resources also restricted our ability to conduct larger-scale experiments, limiting our capacity to explore the full potential of our proposed methodology. Specifically, questions remain regarding how our methodology would perform under systematically optimized hyperparameters and with extended training epochs. Addressing these could provide deeper insights into the scalability and robustness of both our proposed methodology and the future of task-specific attention mechanisms.

\subsection{Future Directions}
Advancements can be made in the engineering of task-specific attention mechanisms. For instance, by incorporating state-of-the-art segmentation and classification techniques, the framework could achieve improved performance and robustness across diverse medical imaging tasks \cite{ghodrati2022automatic} \cite{huang2023self}. While this work focuses on chest X-rays, the MXA block and KD framework can also be adapted for other imaging modalities, such as CT scans, MRIs, or ultrasounds. These modalities often contain richer spatial and temporal information, which could further benefit from the integration of task-specific attention mechanisms.

\section{Conclusion}
We have presented a task-aware attention framework that bridges the gap between generic vision transformers and the demands of clinical X-ray diagnosis. By embedding a lightweight Medical X-ray Attention (MXA) block alongside a calibrated DenseNet-121 teacher, our EfficientViT-MultiLabel model achieves a remarkable +0.19 absolute AUC gain on CheXpert—while preserving inference cost. These results underscore the power of domain-specific attention and resource-efficient distillation. In practice, our approach can be deployed on standard hospital workstations to accelerate and augment radiology workflows. We believe this work lays a versatile blueprint for marrying transformer expressivity with practical deployment constraints in healthcare and beyond.

\begin{ack}
We would like to express our gratitude to Sam Lin for their help in creating Figure 1, along with a research poster, which greatly enhanced the clarity and presentation of our work. We also thank Lambda Lab support for their guidance in overcoming issues with our cloud environment. Additionally, our research was supported with Cloud TPUs from Google's TPU Research Cloud (TRC).
\end{ack}

{
\bibliographystyle{unsrt}
\bibliography{references}
}


\appendix
\counterwithin{table}{section}
\counterwithin{figure}{section}
\renewcommand{\thetable}{\Alph{section}.\arabic{table}}
\renewcommand{\thefigure}{\Alph{section}.\arabic{figure}}

\FloatBarrier       
\clearpage

\section{Formal definition of the U-1 label protocol}
\label{app:u1_formal}
Formally, for each pathology label
\[
y_c=\begin{cases}
1,&\text{label is 1 or -1 (uncertain)},\\
0,&\text{otherwise}.
\end{cases}
\]
Thus, each image is paired with a multi-label target \(y \in \{0,1\}^{C}\), where \(C = 14\) corresponds to the number of binary pathology indicators provided per CXR in our CheXpert training set.

\section{Formal definition of binary cross‑entropy with logits (BCEWL)}
\label{app:bcewl}
\[
\mathcal{L}_{\text{BCEWL}}(o,y)=
-\!\!\sum_{c=1}^{C}\!\bigl[
y_c\log\sigma(o_c)+(1-y_c)\log\bigl(1-\sigma(o_c)\bigr)\bigr],
\]
where $\sigma$ is the sigmoid.

\section{Formal definition of ROI}
\label{app:def_roi}
Given an input feature map $F \in \mathbb{R}^{B \times C \times H \times W}$, where $B$ is the batch size, $C$ the number of channels, and $H,W$ the height
and width, our ROI‑pooling mechanism selects critical regions via a region
predictor, formally defined as
\[
\mathrm{ROI}_i = \bigl[x_1,\,y_1,\,x_2,\,y_2\bigr],
\qquad i = 1,\ldots,B,
\]
where $\bigl[x_1,y_1\bigr]$ and $\bigl[x_2,y_2\bigr]$ are the normalized
top‑left and bottom‑right coordinates of the ROI for the $i$‑th image. These coordinates are generated by a lightweight convolutional network conditioned on the feature map. The bounding-box region predictor is trained jointly with the classification objective in an end-to-end fashion: during the forward pass, each predicted ROI is used to crop and rescale the input feature map, and the resulting ROI features flow into the rest of the network for classification. The gradients from the multi-label classification loss then back-propagate through the ROI predictor as well, optimizing both the bounding boxes and the classification parameters simultaneously. Consequently, the network learns ROIs that best localize relevant abnormalities for improved multi-pathology detection.

\section{Formal definition of CBAM-style attention}
\label{app:cbam_attention}
For a given pooled feature map $F_{\text{pooled}}\in\mathbb{R}^{C\times H\times W}$, channel attention is computed as
\[
M_C
\;=\;
\sigma\!\Bigl(
  W_2\,\delta\!\bigl(W_1\,\operatorname{GAP}(F_{\text{pooled}})\bigr)
  + W_2\,\delta\!\bigl(W_1\,\operatorname{GMP}(F_{\text{pooled}})\bigr)
\Bigr),
\]
where $W_1\in\mathbb{R}^{C\times C/r}$ and $W_2\in\mathbb{R}^{C/r\times C}$ are learnable weight matrices, $\delta$ denotes the ReLU activation \cite{nair2010relu}, $\sigma$ is the sigmoid function \cite{han1995sigmoid}, and $r$ is the reduction ratio controlling the dimensionality of the MLP.  
The resulting attention map $M_C\in\mathbb{R}^{C}$ is applied element‑wise to the feature map:
\[
F_{\text{chan}}\;=\;M_C\odot F_{\text{pooled}}.
\]

Next, spatial attention identifies the most critical regions within each channel‑refined feature map \cite{woo2018cbam}. Using $F_{\text{chan}}$, we pool along the channel dimension with max and average pooling to obtain
\[
F_{\text{spatial}}
\;=\;
\bigl[
  \operatorname{MaxPool}(F_{\text{chan}}),\;
  \operatorname{AvgPool}(F_{\text{chan}})
\bigr],
\qquad
F_{\text{spatial}}\in\mathbb{R}^{2\times H\times W}.
\]
These descriptors are concatenated and passed through a convolutional layer to produce the spatial attention map
\[
M_S \;=\; \sigma\!\bigl(\operatorname{Conv2D}(F_{\text{spatial}})\bigr),
\qquad
M_S\in\mathbb{R}^{H\times W}.
\]
The spatial attention map is then applied element‑wise:
\[
F_{\text{spat}} \;=\; M_S\odot F_{\text{chan}}, 
\qquad
F_{\text{CBAM}} \;=\; F_{\text{spat}}.
\]

\section{Formal definition of soft knowledge distillation term}
\label{app:distill_term}
Let $O^{\text{t}}\in\mathbb{R}^{18}$ denote the teacher logits and the student outputs logits be \(O^{\mathrm{s}}\). Let the adapted teacher probabilities be represented as \(p^{\mathrm{t}}=\sigma(O^{\mathrm{t}})\). Thus, our soft distillation is defined as
\[
\mathcal{L}_{\mathrm{KD}}^{\mathrm{soft}}
\;=\;
\frac{1}{C}\sum_{j=1}^{C}
w_{j}\,
\operatorname{BCEWithLogitsLoss}\!\bigl(O^{\mathrm{s}}_{j},\,p^{\mathrm{t}}_{j}\bigr),
\]
with \(C=14\) labels.  

\FloatBarrier
\clearpage

\section{Qualitative MXA ROI examples and patient-level metadata}
\label{app:mxa_qual}

\begin{figure}[h!]
  \centering
  \captionsetup[subfigure]{justification=centering} 
  \foreach \i in {1,...,8}{%
    \begin{subfigure}[b]{0.24\linewidth}
      \centering
      \includegraphics[width=\linewidth]{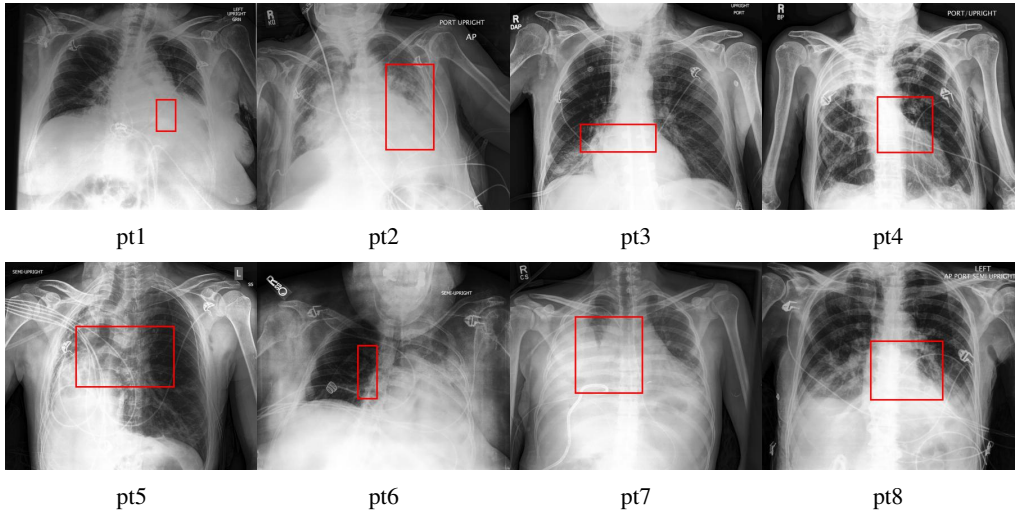}
      \caption*{pt\i}        
    \end{subfigure}%
    \ifnum\i=4\par\smallskip\fi
  }
  \caption{Demonstration of the MXA block. MXA consistently highlights clinically relevant regions across eight patients, confirming its ability to localize subtle abnormalities. Each panel (pt1–pt8) shows the region of interest automatically pooled after an initial inference on chest X‑rays. Red boxes mark MXA‑predicted ROIs. Additional pathology metadata appear in Table~\ref{tab:mxa_metadata}.}
  \label{fig:mxa_demo}
\end{figure}
\begin{table}[h!]
\centering
\caption{Metadata for the 8 MXA test figures}
\label{tab:mxa_metadata}
\resizebox{\textwidth}{!}{%
\begin{tabular}{lcccccccccccccc}
    \toprule
    \textbf{Patient} & \textbf{NF} & \textbf{ECM} & \textbf{CM} & \textbf{LO} & \textbf{LL} & \textbf{ED} & \textbf{CON} & \textbf{PNA} & \textbf{ATL} & \textbf{PTX} & \textbf{PE} & \textbf{PO} & \textbf{FX} & \textbf{SD} \\
    \midrule
    pt1 & 0 & 1 & 1 & 1 & 0 & 0 & 1 & 1 & 1 & 0 & 1 & 0 & 0 & 0 \\
    pt2 & 0 & 1 & 1 & 1 & 0 & 1 & 1 & 1 & 1 & 0 & 1 & 0 & 0 & 0 \\
    pt3 & 0 & 0 & 0 & 1 & 0 & 0 & 1 & 1 & 0 & 0 & 0 & 0 & 0 & 1 \\
    pt4 & 0 & 0 & 0 & 1 & 0 & 0 & 1 & 1 & 1 & 0 & 1 & 0 & 0 & 0 \\
    pt5 & 0 & 0 & 0 & 1 & 0 & 0 & 1 & 1 & 1 & 0 & 1 & 1 & 0 & 1 \\
    pt6 & 0 & 1 & 1 & 1 & 0 & 0 & 1 & 1 & 1 & 0 & 1 & 0 & 0 & 0 \\
    pt7 & 0 & 1 & 1 & 1 & 0 & 0 & 1 & 1 & 1 & 0 & 1 & 0 & 0 & 1 \\
    pt8 & 0 & 0 & 0 & 1 & 0 & 1 & 1 & 1 & 1 & 0 & 1 & 0 & 0 & 0 \\
    \bottomrule
\end{tabular}%
}
{\footnotesize
Abbreviations: 
NF = No Finding, ECM = Enlarged Cardiomediastinum, CM = Cardiomegaly, LO = Lung Opacity, LL = Lung Lesion, ED = Edema, CON = Consolidation, PNA = Pneumonia, ATL = Atelectasis, PTX = Pneumothorax, PE = Pleural Effusion, PO = Pleural Other, FX = Fracture, SD = Support Devices.
}
\end{table}

\FloatBarrier 
\clearpage 

\section{Additional Qualitative MXA examples on CXRs}
\begin{figure*}[h]
  \centering
  \includegraphics[width=\linewidth]{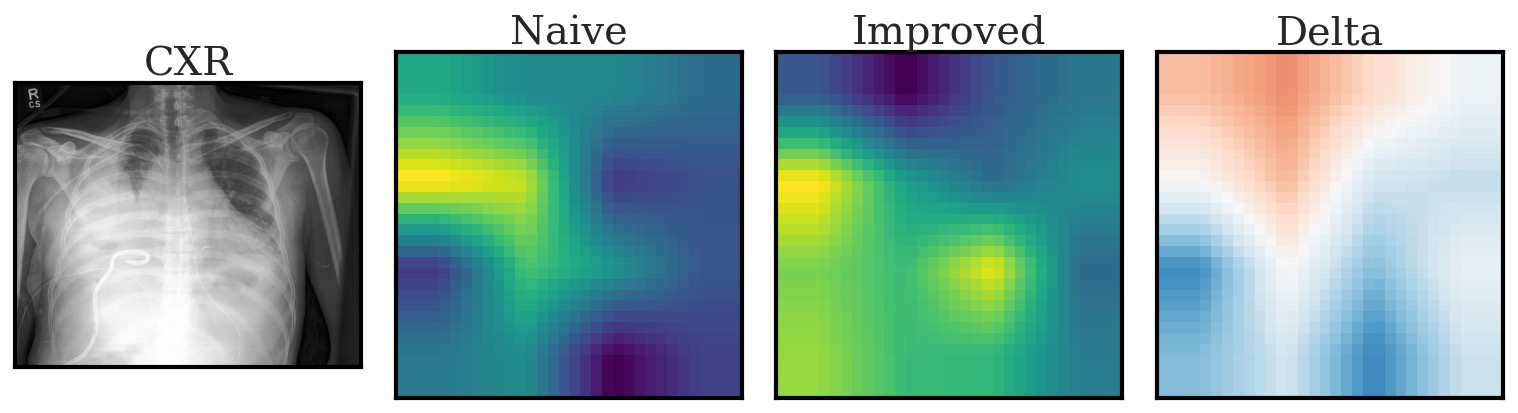}
  \includegraphics[width=\linewidth]{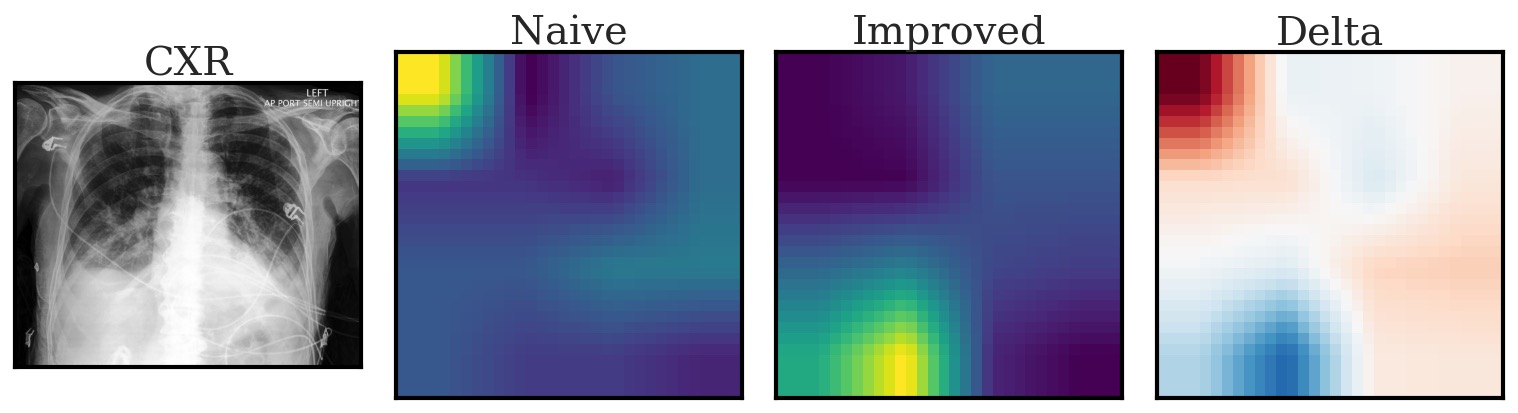}
  \includegraphics[width=\linewidth]{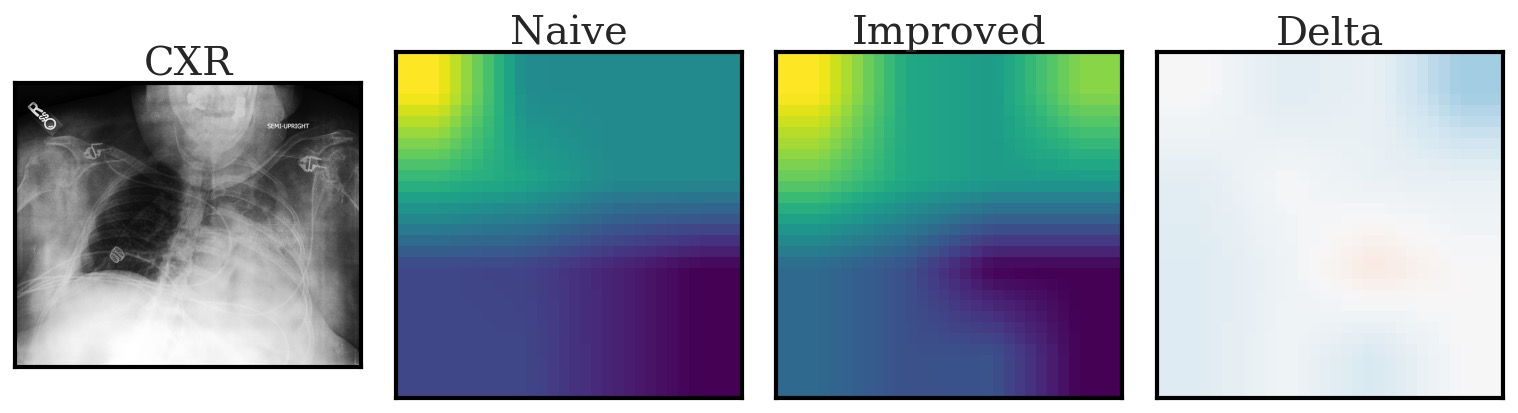}
  \caption{Additional attention map visualizations after 25 training epochs are shown. Across CXRs with pneumonia and diverse acquisition geometries the MXA block  consistently (i) reduces attention on bony anatomy and monitoring
  hardware, and (ii) reallocates it toward pleural
  regions containing abnormalities, corroborating the quantitative
  AUC gains reported in Tab.~\ref{tab:overall_performance}. Bright yellow in the Naive/Improved panels = high attention; dark purple = low attention. The Delta panel shows the difference in attention: red indicates regions where MXA attends \textit{less} than naive MHSA, blue where it attends \textit{more}, and white means no change.} 
  \label{fig:qualitative_appendix}
\end{figure*}

\FloatBarrier 
\clearpage 

\section{Multi-label EfficientViT design space}
\label{app:design_space}

Table~\ref{tab:efficientvit} lists the six principal variants of our multi-label EfficientViT architectures we defined. Each model family member differs in its stage-wise width ($C_i$), depth ($L_i$), and number of heads ($H_i$). Specifically, we break the network into three stages, and each stage’s dimensions are adjusted to control the overall capacity and computational cost. Resource limits restricted our experiments to the M5 configuration, which offers the best parameter–FLOP trade-off within that budget. Future work will explore the remaining variants.

\begin{table}[h!]
  \centering
  \caption{Multi-label EfficientViT architecture variants}
  \label{tab:efficientvit}
  \begin{tabular}{lccc}
\toprule
    \textbf{Model} & $\{C_1,C_2,C_3\}$ & $\{L_1,L_2,L_3\}$ & $\{H_1,H_2,H_3\}$ \\
    \midrule
    EfficientViT-MultiLabel-M0 & \{64, 128, 192\} & \{1, 2, 3\} & \{4, 4, 4\} \\
    EfficientViT-MultiLabel-M1 & \{128, 144, 192\} & \{1, 2, 3\} & \{2, 3, 3\} \\
    EfficientViT-MultiLabel-M2 & \{128, 192, 224\} & \{1, 2, 3\} & \{4, 3, 2\} \\
    EfficientViT-MultiLabel-M3 & \{128, 240, 320\} & \{1, 2, 3\} & \{4, 3, 4\} \\
    EfficientViT-MultiLabel-M4 & \{128, 256, 384\} & \{1, 2, 3\} & \{4, 4, 4\} \\
    EfficientViT-MultiLabel-M5 & \{192, 288, 384\} & \{1, 3, 4\} & \{3, 3, 4\} \\
    \bottomrule
  \end{tabular}
\end{table}

\FloatBarrier       
\clearpage

\section{Additional quantitative plots}
\label{app:extra_plots}

\begin{figure}[h!]
  \centering
  \includegraphics[width=0.5\linewidth]{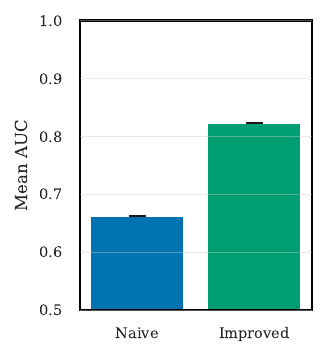}
  \caption{Mean AUC over three validation runs for baseline and proposed models; error bars denote 95\% confidence intervals ($\pm2\sigma$). Across three runs, MXA + KD improves mean AUC drastically over the baseline, demonstrating robustness to random seed.}
  \label{fig:mean_auc_ci}
\end{figure}

\begin{figure}[h!]
  \centering
  \includegraphics[width=\linewidth]{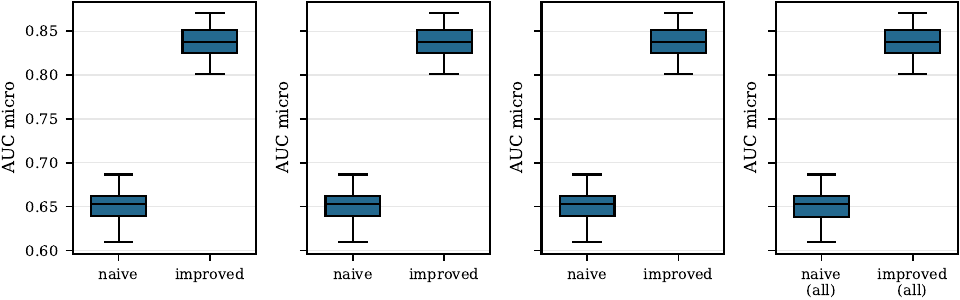}
  \caption{Box-plot comparison of per-epoch AUC for three runs
           (baseline vs.\ proposed). Every epoch shows higher and less variable AUC for MXA + KD, indicating both greater accuracy and consistency than the baseline.}
  \label{fig:boxplot_auc_runs}
\end{figure}

\FloatBarrier 
\clearpage 

\section{Reported compute resources}
\label{app:rcr}

\begin{table}[h!]
\centering
\caption{Compute Resources and Usage}
\begin{tabular}{ll}
\toprule
\textbf{Resource} & \textbf{Specification} \\
\midrule
\multicolumn{2}{l}{\textbf{Lambda Labs Node}} \\
\quad GPU & 1$\times$ NVIDIA H100 (80~GB PCIe) \\
\quad CPU & 26 vCPUs \\
\quad Memory & 200~GiB RAM \\
\quad Storage & 1~TiB SSD \\
\quad Number of experiments & $\sim$120 \\
\quad Avg. compute time per experiment & $\sim$10 hours \\
\quad Total compute time & $\sim$1,200 hours \\
\midrule
\multicolumn{2}{l}{\textbf{Google TPU}} \\
\quad TPU & 1$\times$ v4-8 \\

\quad Number of experiments & $\sim$1 \\
\quad Avg. compute time per experiment & $\sim$50 hours \\
\quad Total compute time & $\sim$ 50 hours 
\end{tabular}
\end{table}

\end{document}